\documentclass[11pt]{article}
\usepackage[final]{acl}
\usepackage{times}
\usepackage{latexsym}
\usepackage[T1]{fontenc}
\usepackage[utf8]{inputenc}
\usepackage{microtype}
\usepackage{inconsolata}
\usepackage{graphicx}
\renewcommand{\cite}{\citep}
\usepackage{graphicx}
\usepackage{amsmath}
\usepackage{amssymb}
\usepackage{booktabs}
\usepackage{geometry}
\usepackage{multirow}
\usepackage{amsmath,amssymb,amsfonts}
\usepackage{algorithmic}
\usepackage{subfigure} 
\usepackage{graphicx}
\usepackage{textcomp}
\usepackage{xcolor}
\title{RetentiveKV: State-Space Memory for Uncertainty-Aware Multimodal KV Cache Eviction}

\author{
    Sihao Liu\textsuperscript{1},
    YuFan Xiong\textsuperscript{2}, 
    Zhonghua Jiang\textsuperscript{2}, 
    Zhaode Wang\textsuperscript{3}, 
    chengfei lv\textsuperscript{3}, 
    Shengyu Zhang\textsuperscript{1 $\dagger$}
    \\
    \textsuperscript{1}Zhejiang University \\
    \textsuperscript{2}Alibaba
    \textsuperscript{3} Shanghai Institute for Advanced Study of Zhejiang University\\
    \url{{lyosihao, Zhonghua Jiang, Shengyu Zhang}@zju.edu.cn} \\
    \url{xiomg@webmail.hzau.edu.cn}\\
    \url{{zhaode.wzd, chengfei.lcf}@taobao.com}
}

\begin{document}
\maketitle
\begin{abstract}
Multimodal Large Language Models face severe challenges in computational efficiency and memory consumption due to the substantial expansion of the visual KV cache when processing long visual contexts. Existing KV cache compression methods typically rely on the "persistence of importance" hypothesis to prune tokens. However, this approach proves fragile in multimodal settings due to two key issues: 1) Visual tokens display "deferred importance," initially exhibiting low salience but becoming pivotal during later decoding, which can lead to premature eviction. 2) Discrete pruning disrupts the inherent spatial continuity of visual cues. To address these challenges, we propose RetentiveKV, an entropy-driven KV cache optimization method that reformulates KV eviction from "discrete context truncation" to "continuous memory evolution" based on State Space Models. Our method leverages information entropy to quantify the information potential of low-attention tokens and integrates tokens scheduled for eviction into a continuous state space through entropy-guided state transitions, enabling their dynamic reactivation when semantic relevance arises during subsequent decoding. Extensive experiments on multimodal benchmarks demonstrate that RetentiveKV achieves 5.0 × KV cache compression and 1.5 × decoding acceleration.
\end{abstract}  
\section{Introduction}
Multimodal Large Language Models (MLLMs) have emerged as the dominant paradigm for understanding and generating cross-modal content. However, the increasing demand for processing long visual contexts and high-resolution inputs results in a substantial expansion of visual tokens and the KV cache. This imposes severe challenges on computational efficiency and memory consumption during inference. Consequently, the efficient optimization of token sequences and the KV cache has become critical for accelerating autoregressive inference in MLLMs.

\begin{figure*}[t]
    \centering
    \subfigure[Temporal Importance Heatmap of Visual Token.]{
        \includegraphics[width=0.48\textwidth]{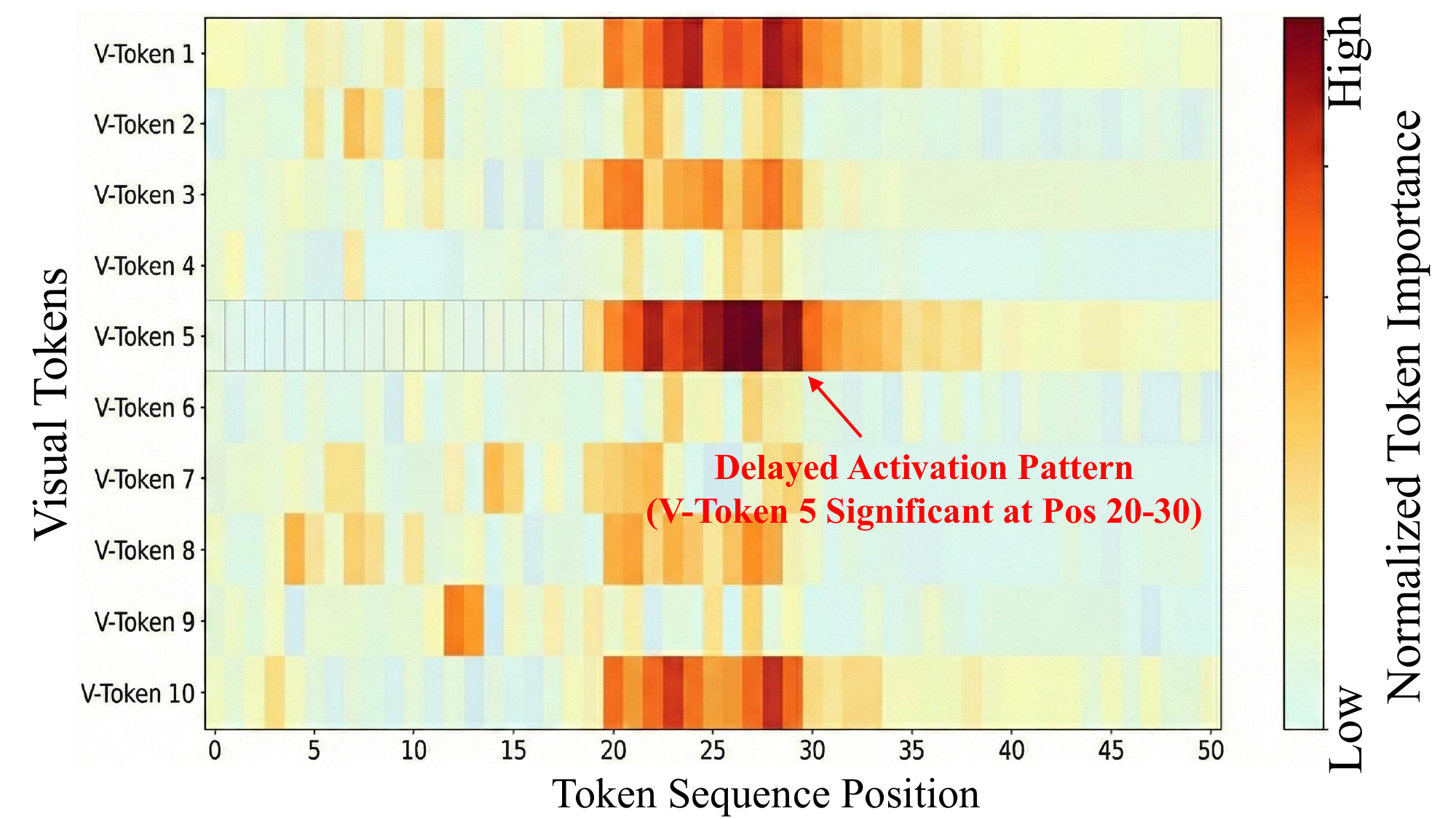}
        \label{fig:latency}
    }
    \hfill
    \subfigure[Impact of Token Eviction on Cross-model Attention.]{
        \includegraphics[width=0.48\textwidth]{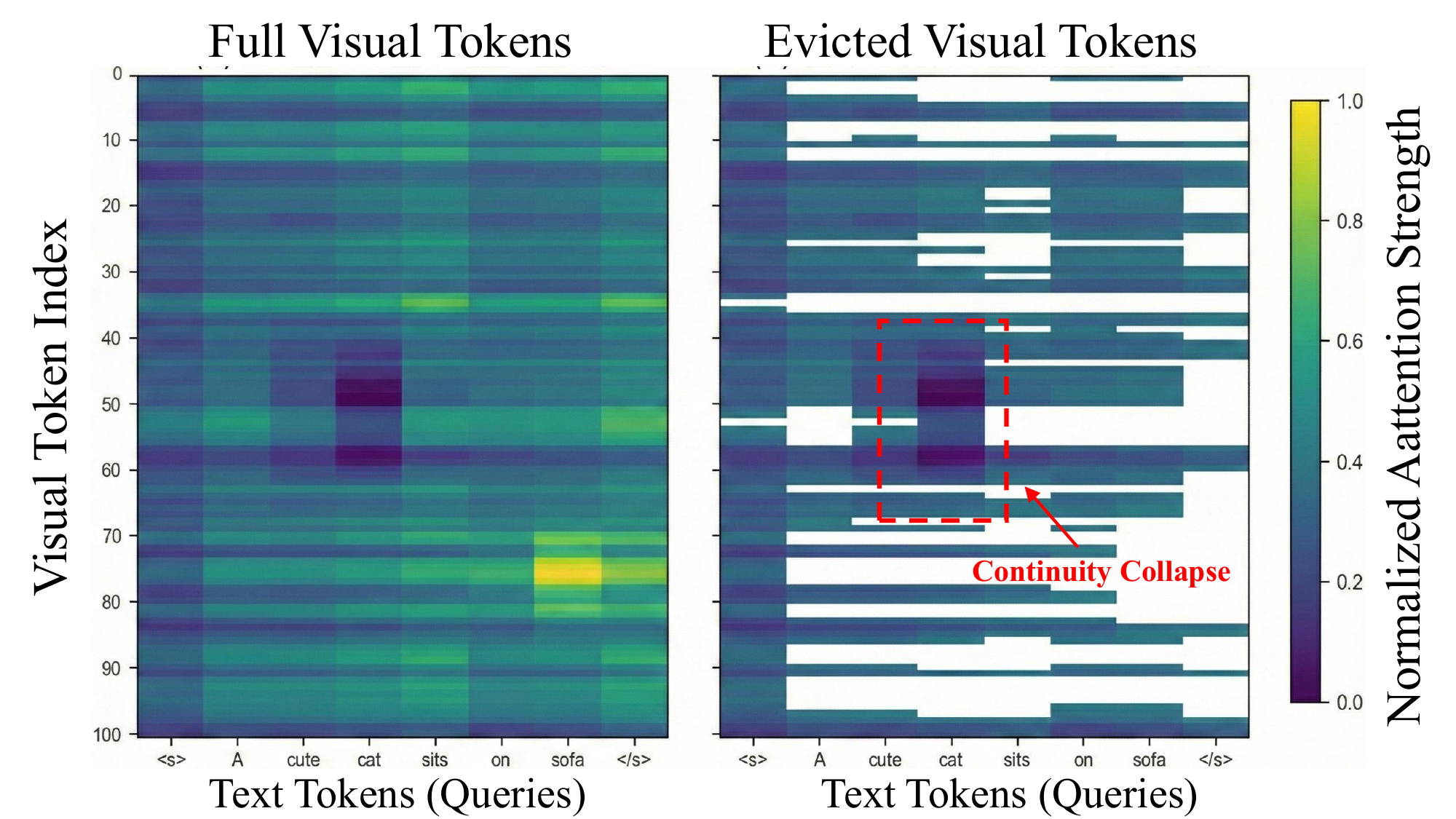}
        \label{fig:sparsity}
    }
    \caption{(a) Visualization of Deferred-Critical Tokens: The presence of delayed activation patterns, where visually salient features remain dormant during early decoding phases, defies the static heavy-hitter assumption. This temporal misalignment triggers erroneous eviction in importance-based pruning. (b) Visualization of Pruning-Induced Spatial Discontinuity: We compare baseline and pruned attention maps. The instability of visual signals triggers excessive pruning (white voids). This fragmentation disrupts the spatial continuity of visual representations.}
    \label{fig:analysis}
\end{figure*}

Prior research in Large Language Models (LLMs) has extensively explored importance-centric KV cache compression methods. These approaches generally leverage token-level importance metrics to selectively evict KV pairs with minimal contributions, thereby mitigating memory consumption and computational complexity. While existing methods demonstrate considerable promise in text-only settings, they prove fragile in multimodal contexts: 1) \textbf{Deferred Importance Matter}: Conventional KV compression relies on the "persistence of importance" hypothesis \cite{h2o,hh1}, assuming that tokens with high initial attention remain critical throughout decoding. This assumption is widely inherited by multimodal methods \cite{fastv,boosting}. However, we observe a distinct "Deferred Importance" phenomenon: visual tokens often exhibit low initial salience but become pivotal at later timesteps. As shown in Figure \ref{fig:latency}, attention distribution in Visual Question Answering (VQA) tasks is temporally non-uniform. Early decoding prioritizes the semantic comprehension of the textual query, while the attention mechanism only pivots toward the cross-modal dependencies of visual tokens when generating key responses requiring specific visual attributes. 2) \textbf{Visual Continuity Collapse}: Visual tokens are characterized by intrinsic spatial continuity and strong inter-patch correlations. As illustrated in Figure \ref{fig:sparsity}, the discrete eviction of KV pairs inevitably fragments these continuous representations. While existing multimodal methods \cite{madakv,acckv} attempt to mitigate this by modulating eviction ratios based on modality preferences, this strategy essentially relies on quantitative reallocation. It fails to fundamentally address the critical issue of spatial discontinuity, as the underlying eviction mechanism remains structure-agnostic.

In this paper, drawing inspiration from State Space Models (SSMs), we transform KV cache eviction from "discrete context truncation" to "continuous memory evolution". Instead of permanently discarding tokens with low attention scores, our approach assimilates them into a continuous state space governed by information entropy. This mechanism preserves visual tokens exhibiting high uncertainty during early decoding stages and updates them continuously as decoding progresses, all while incurring only $O(1)$ memory overhead. Furthermore, by replacing the pair-token softmax kernels of standard attention with recursive convolution kernels derived from state transitions, SSMs inherently maintain the spatial continuity of visual cues, mitigating visual continuity collapse.

To this end, we propose RetentiveKV, a framework that reformulates KV cache eviction via entropy-driven SSMs. Specifically, RetentiveKV comprises three innovations for incorporating evicted information into a dynamic state space: 1) Entropy-guided KV Retention Estimator: Traditional KV eviction typically adheres to the "persistence of importance" principle, which assumes that tokens exhibiting high attention scores in current steps will remain critical for subsequent decoding. In contrast, our method incorporates an additional "prospective uncertainty" principle, designed to estimate the "undetermined cross-modal potential" of tokens exhibiting low attention scores. We leverage the entropy of the attention distribution from textual to visual tokens to assess this potential. High entropy indicates that a textual token attends diffusely across visual tokens, implying that the visual context holds a high degree of uncertainty and thus potential relevance for subsequent decoding steps. 2) Entropy-Guided State Transition: This mechanism reformulates the discrete KV eviction process into a continuous state evolution, which leverages selective state to approximate the per-token interactions of self-attention. Compared to weighted merging strategies, this formulation preserves the ability to selectively attend to specific historical segments while naturally incorporating positional and spatial information through recursive state updates. 3) State Modeling and Retrieval: Inspired by the hierarchical nature of human memory \cite{memory}, we propose a dual-state architecture to resolve the conflict between visual continuity and semantic discreteness. Specifically, we design Visual-Dominant States to preserve the spatial topology of visual patches, and Recall-Oriented States to capture long-term semantic dependencies. These states are continuously updated via entropy-guided transitions. During inference, a Query-Conditioned Retrieval mechanism dynamically queries these states, retrieving and reactivating "deferred-critical" information only when semantic relevance to the current query is detected.

Our contributions are summarized as follows: 

1. We reformulate KV eviction as continuous state evolution, assimilating 'deferred-critical' tokens with undetermined potential into a continuous state representation to preserve spatial continuity.

2. We propose RetentiveKV, which leverages attention entropy to quantify future relevance and compresses evicted tokens into selective states, enabling their dynamic reactivation for decoding.

3. We conduct extensive experiments on multimodal benchmarks, demonstrating that RetentiveKV achieves substantial reductions in memory consumption and computational overhead.

\section{Related Work}
\subsection{KV Cache Eviction}
 The autoregressive decoding mechanism of LLMs necessitates caching Key and Value (KV) states to avoid redundant re-computation. However, the linear growth of the KV cache with respect to sequence length imposes a significant memory bottleneck. To mitigate this, importance-centric eviction strategies \cite{flowmm,purekv} leverage the "heavy-hitter" phenomenon, observing that a small subset of tokens accumulates the majority of attention mass. While these seminal works effectively exploit attention sparsity to prune redundant KV pairs, subsequent research has refined the definition of token importance. For instance, StreamingLLM \cite{stream} uncovers the "attention sink" phenomenon, advocating for the persistence of initial tokens alongside the most recent local context. Extending the scope to MLLMs, the memory bottleneck is further compounded by the expansion of visual tokens generated from high-resolution images and long-form videos. Furthermore, SnapKV \cite{snapkv} introduces a voting mechanism that clusters key context chunks exhibiting high responsiveness to the prompt. In this domain, FastV \cite{fastv} investigates the layer-wise distribution of visual importance, uncovering the phenomenon of "visual attention inefficiency". It implements a depth-adaptive eviction policy that discards visual KV states in deeper layers. Similarly, LOOK-M \cite{lookm} introduces a modality-aware compression framework, which utilizes pivotal merging to aggregate spatially correlated visual tokens into compact representations. Meda \cite{meda} dynamically allocates KV cache budgets with information entropy across different transformer layers. SAINT \cite{SAINT} employs a graph-theoretic similarity metric to prune redundant visual tokens within early network layers.
\subsection{State Space Models}
State Space Models originate from classical control theory and signal processing, where they model continuous systems through latent state evolution. However, applying vanilla SSMs to deep learning was historically hindered by prohibitive computational costs and the difficulty of handling long-range dependencies. HiPPO (High-Order Polynomial Projection Operators) \cite{hippo} pioneered a theoretical framework to solve this by projecting continuous signals onto orthogonal polynomial bases, mathematically guaranteeing the optimal retention of history. Building on this, S4 \cite{s4} introduced a parameterization of the state matrix $\mathbf{A}$ as a diagonal plus low-rank structure. This innovation allowed the recurrence to be computed efficiently via parallel scans in the GPU, reducing the complexity from quadratic $O(L^2)$ to linear or log-linear $O(L \log L)$. The most significant recent advancement is the selective SSM \cite{mamba,retnet}. Unlike prior models, where parameters are time-invariant, selective SSM makes these parameters input-adaptive. This allows the model to selectively propagate or forget information based on the current token, effectively solving the "selection mechanism" problem that static SSMs faced. Fundamentally, an SSM maps an input sequence $x_t \in \mathbb{R}$ to an output sequence $y_t \in \mathbb{R}$ through an implicit latent state $h_t \in \mathbb{R}^N$. This process can be formulated as:
$$
\begin{aligned} h_{t+1} &= \mathbf{A}h_t + \mathbf{B}x_t, \\ 
y_t &= \mathbf{C}h_t,
\end{aligned}
$$
where $\mathbf{A}\in \mathbb{R}^{N \times N}$ represents the state evolution matrix, while $\mathbf{B} \in \mathbb{R}^{N \times 1}$ and $\mathbf{C} \in \mathbb{R}^{1 \times N}$ are projection parameters for the input and state. 

\section{Method}

\subsection{SSM-based KV Cache Retention}
Standard Transformers rely on the Scaled Dot-Product Attention mechanism, defined as:

\begin{equation}
\text{Atten}(Q, K, V) = \text{Softmax}\left(\frac{QK^\top}{\sqrt{d}}\right)V.  
\end{equation}

This requires maintaining a KV cache that grows linearly with sequence length $L$, resulting in $O(L)$ memory complexity and $O(L^2)$ computational complexity during decoding.

In this work, we leverage the dual computation paradigm of SSM, which reformulates the attention mechanism as SSM. By removing the non-linear Softmax operation and utilizing the associative property of matrix multiplication, the output computation can be rewritten as:
\begin{equation}
  \text{SSM}(Q, K, V) = Q_t \left( \sum_{i=1}^{t} \gamma^{t-i} (K_i^\top V_i) \right),  
\end{equation}
here, we introduce a decay factor $\gamma \in (0, 1]$ (analogous to the diagonal of the matrix $\bar{\mathbf{A}}$ in SSMs) to modulate the retention of historical information. We can thus define a matrix-valued hidden state $S_t \in \mathbb{R}^{d \times d}$, which serves as a compressed representation of the KV cache. The recurrence rule for RetentiveKV is formally defined as:

\begin{equation}
    \begin{aligned} h_{t+1} = \gamma h_{t} + K_t^\top V_t, \\ O_t = Q_t S_t. \end{aligned}
\end{equation}

In this formulation, $S_t$ effectively absorbs the visual tokens $(K_t, V_t)$ into the state space. Unlike the standard KV cache, which appends tokens to a growing list, this formulation updates the state in-place. This ensures that the memory footprint for the visual context remains constant, regardless of the visual resolution or sequence length.

\subsection{Overview of RetentiveKV}
The RetentiveKV framework consists of three important components to incorporate evicted tokens into entropy-driven state spaces: 1) Entropy-Guided KV Retention Estimator, which leverages entropy-based measures to quantify the uncertainty and prospective relevance of KV pairs for future decoding; 2)  Entropy-Guided State Transition, which absorbs evicted KV cache into a modality-specific state space and leverages token-level entropy variations to modulate retention and decay for state evolution. 3) Query-Conditioned State Retrieval, which is responsible for selectively recalling task-relevant multimodal information from the state space during autoregressive decoding.

\subsection{Observation}
\begin{figure}[h]
    \centering
    \includegraphics[width=\linewidth]{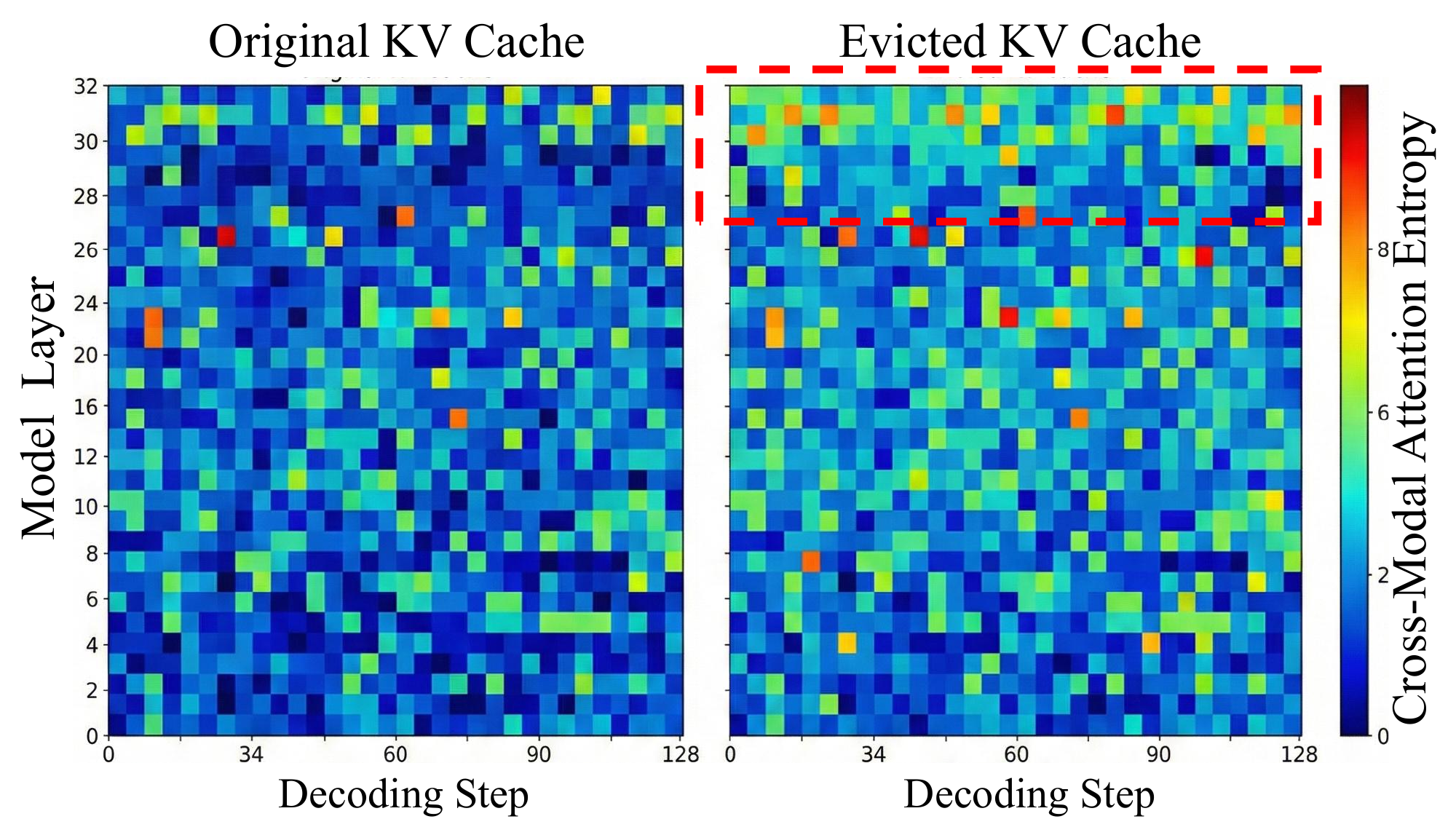}
    \caption{Entropy Shifts under KV Eviction.}
    
    \label{fig:entropy}
\end{figure}

Recent studies \cite{earl,uncomp} in the textual modality have demonstrated that information entropy serves as an important indicator for token compression. Tokens with higher entropy typically correspond to critical decision points in autoregressive decoding, reflecting peaks in contextual shifts that influence the likelihood of subsequent states. Building upon these insights, we investigate the applicability of information entropy as a guiding metric for multimodal token compression. In the MLLMs, the entropy dynamics are complicated by the cross-modal interactions between visual and textual modalities. Therefore, we focus on the information entropy of the attention distribution from textual tokens to visual tokens, which we refer to as \textbf{Cross-Modal Attention Entropy}. As shown in Figure \ref{fig:entropy}, empirical analysis reveals that importance-centric KV eviction induces a layer-dependent increase in cross-modal attention entropy. Some layers exhibit anomalously elevated entropy compared to their non-evicted counterparts, indicating that these layers experience heightened uncertainty regarding cross-modal alignment after KV cache eviction. The layers exhibiting significant increases in cross-modal attention entropy are mainly located in the middle-to-upper depths of the model. These layers are widely regarded as playing a critical role in abstract semantic modeling and cross-modal alignment. These observations motivate the adoption of cross-modal attention entropy as an important metric for multimodal KV cache compression.

\begin{figure*}
    \centering
    \includegraphics[width=\linewidth]{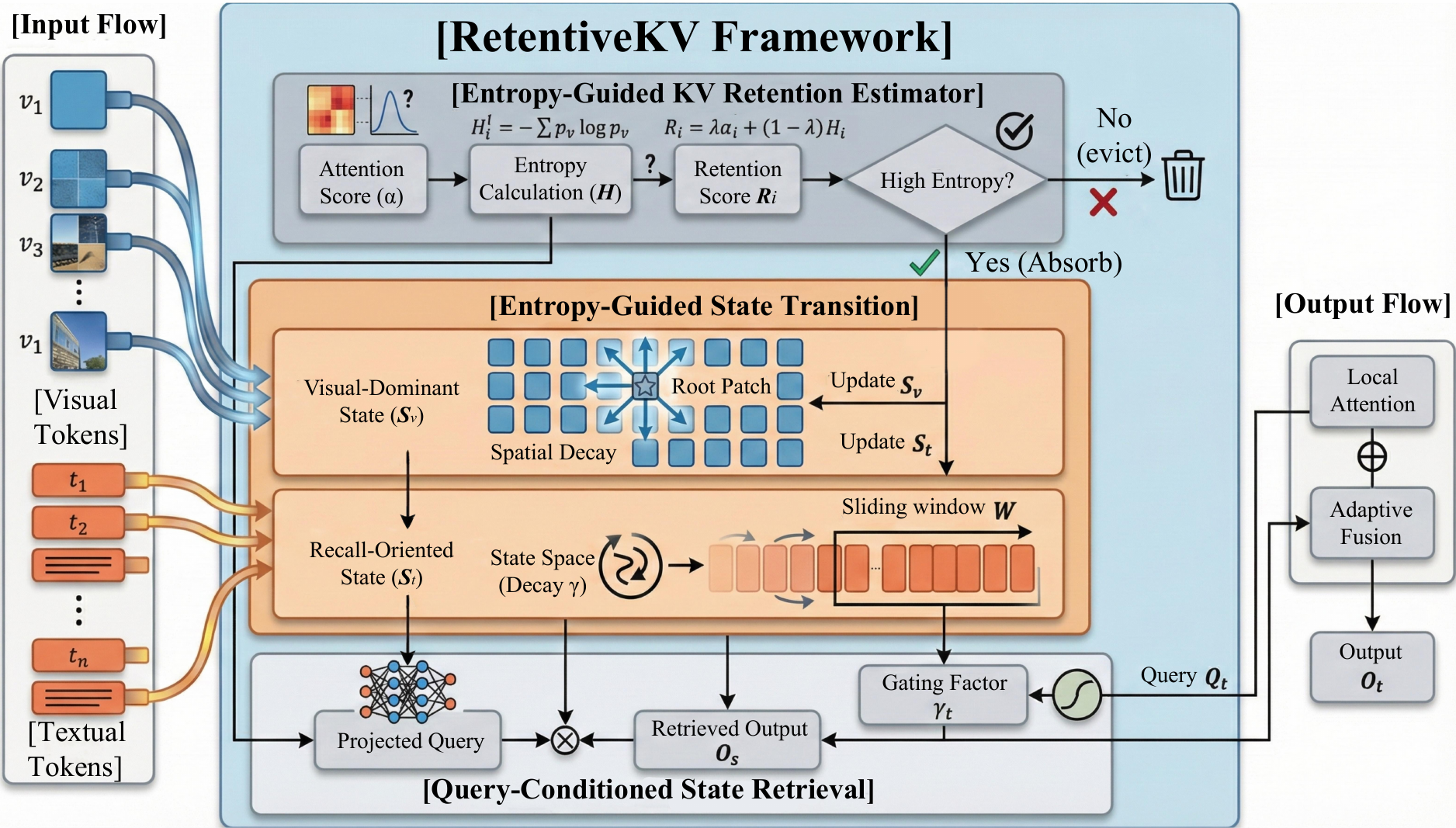}
    \caption{Overview of the proposed RetentiveKV framework. The architecture coordinates efficient long-context multimodal reasoning through three core components: (1) An Entropy-Guided KV Retention Estimator that identifies deferred-critical tokens by analyzing layer-wise entropy shifts ($H_t$) and accumulated attention ($\alpha_t$); (2) An Entropy-Guided State Transition mechanism that absorbs evicted KV pairs into modality-adaptive state space; and (3) A Query-Conditioned State Retrieval module that dynamically fuses retrieved long-term context ($O_s$) with local attention outputs via a learnable gating factor ($\gamma_t$) during autoregressive decoding.}
    \label{fig:overall}
\end{figure*}

\subsection{Entropy-Guided KV Retention Estimator}
Entropy-Guided KV Retention Estimator introduces the cross-modal attention entropy to measure the distributional uncertainty of attention from textual tokens to visual tokens. Let $\alpha_t^{l, i}$ denotes the standard attention score for token $i$ at decoding step $t$ for the $l$-th layer, $p_v(\cdot)$ represents the cross-modal attention scores selected from the $\alpha_t^{l, i}$. The cross-modal attention entropy is defined by the Shannon entropy over the attention distribution. Formally: 
\begin{equation}
    H_t^{l,i} = -p_v(\alpha_t^{l,i}) \cdot \log p_v(\alpha_t^{l,i}).
\end{equation}

We define a retention score $R^{l,i}$ to decide whether the KV pair of token $i$ should be retained in the state at layer $l$. Our scoring function integrates both the immediate importance and the prospective uncertainty. Formally, the retention score is calculated as a weighted combination of the standardized attention score and the cross-modal attention entropy:
\begin{equation}
    R_t^{l,i} = \lambda \alpha_t^{l,i} + (1-\lambda)  H_t^{l,i},
\end{equation}
where $\lambda \in [0,1]$ balances the contribution of immediate importance and prospective utility. KV pairs with $R_t$ above a retention threshold $\tau$ are absorbed into the continuous state space.

\subsection{Entropy-Guided State Transition}
The objective of Entropy-Guided State Transition is to integrate evicted KV pairs into a continuous state space and govern the evolution of the state space. Differing from standard SSMs with a fixed transition matrix, our method introduces cross-modal attention entropy as a dynamic transition coefficient. Formally, we define the continuous state at decoding step $t$ as $\mathbf{S}_t$. The state transition is formulated as a recursive update equation:

\begin{equation}
    \mathbf{S}_t= \mathbf{H}_t \odot \mathbf{S}_{t-1} +  \mathbf{A}_t \odot (\mathbf{k}_t^\top \mathbf{v}_t),
\end{equation}
where $\mathbf{A}$ denotes the absorption matrix that controls the injection of the current evicted KV pair, modulated by the normalized accumulated attention of the current tokens, and $\mathbf{H}$ denotes the retention matrix that determines how much of the accumulated state is decayed, modulated by the token-level information entropy. Formally, let the normalized accumulated attention score and the sigmoid-transformed cross-modal attention entropy of token $i$ be $\alpha_i$ and $\sigma(H_i)$, respectively. We define:

\begin{equation}
    \mathbf{H}_t[n,m]=\left\{\begin{array}{ll}
\sigma(H_t)^{n-m}, & n \geq m \\ 
0, & n<m 
\end{array}\right. ,
\end{equation}

\begin{equation}
\mathbf{A}_t[n,m]=\left\{\begin{array}{ll}
\alpha_t^{n-m}, & n \geq m \\ 
0, & n<m 
\end{array}\right. .
\end{equation}

The Entropy-Guided State Transition mechanism defined above serves as the foundational KV retention kernel for RetentiveKV. While the above equation defines the general form of state evolution, its deployment varies across different phases of autoregressive decoding.

\subsection{Modality-Agnostic Initial State Modeling}
The core challenge during the prefill stage lies in the parallel processing of high-resolution visual tokens and long-sequence textual instructions, which requires the model to manage a large-scale multimodal KV cache concurrently. In the prefill stage, we adopt the following dual-state strategies:
\subsubsection{Visual-Dominant State ($\mathbf{S}_{V}$)} Visual tokens possess an inherent two-dimensional spatial topology and continuity. To preserve these geometric priors, we model the visual-dominant state as a spatially-aware State Space. For each input image $\mathcal{I}$, we maintain an independent state $\mathbf{S}_{V}^{\mathcal{I}}$, which is decomposed into two orthogonal sub-states representing horizontal and vertical scanning directions. The initialization of the state space is conditioned on the root patch  $\mathbf{p}_{x_m,y_m}$ with the maximum accumulated attention score.  For a patch located at grid coordinates $(x_n, y_n)$ being evicted from the discrete cache, its integration into the continuous state is modulated by its spatial displacement from the root patch. We define the update of the visual-dominant state as:
\begin{equation}
\begin{aligned}
\mathbf{S}_{V,t}^{\mathcal{I}}
&= \mathbf{H}_t^{|x_n-x_m|+|y_n-y_m|} \odot \mathbf{S}_{V,t-1}^{\mathcal{I}} \\
&\quad + \mathbf{A}_t^{|x_n-x_m|+|y_n-y_m|} \odot (\mathbf{k}_t^\top \mathbf{v}_t).
\end{aligned}
\end{equation}

\subsubsection{Recall-Oriented State ($\mathbf{S}_{T}$)} For the textual modality, we maintain a fixed-length working sliding window of size $W$, where the most recent $W$ key-value pairs are preserved in full-precision. These instantaneous memories act as high-resolution semantic anchors, preventing the accumulation of approximation errors induced by early-stage eviction. For tokens that are outside the sliding window or exiting the sliding window during decoding, we identify the textual tokens that are not classified as "Heavy-Hitters" \cite{h2o} but exhibit high attention entropy and are integrated into the Recall-Oriented State.

\subsection{Query-Conditioned State Retrieval}
To effectively leverage the information retained in the continuous state space during decoding, we introduce a query-conditioned state retrieval mechanism, which conditions on the current query states to selectively re-inject features from the visual-dominant state $(\mathbf{S}_{\text{V}})$ and the recall-oriented state $\mathbf{S}_{\text{T}}$ into the attention computation. 

At each decoding step $t$, given the current query vector $\mathbf{q}_t$, we retrieve the contextual features from the dual state spaces by projecting the query into the continuous state space. The retrieved value representation $\mathbf{v}^{(t)}$ is defined as:
\begin{equation}
    \mathbf{O}_{\mathbf{S}}^{(t)} = \text{Norm}(\mathbf{q}_t \mathbf{S}_{V}^{(t-1)} + \gamma_t \cdot\mathbf{q}_t \mathbf{S}_{T}^{(t-1)}),
\end{equation}
where $\mathbf{S}_{V}^{(t-1)}$ denotes the visual-dominant state constructed by concatenating the instance-specific visual states ${\mathbf{S}_{V}^{\mathcal{I}}}$ corresponding to input images, $\text{Norm}(\cdot)$ denotes Layer Normalization, which is applied to stabilize the numerical distribution of the fused multimodal features. We introduce an activation gating factor $\gamma_t$, which adaptively controls the contribution of the retrieved state information. Formally, we define the activation gating factor as:
\begin{equation}
    \gamma_t = \sigma \left( W_r \cdot H_t + b_r \right),
\end{equation}
where $H_t$ denotes the average cross-modal attention entropy at decoding step $t$, and $W_r$ and $b_r$ are learnable parameters. The sigmoid function $\sigma(\cdot)$ constrains $\gamma_t \in (0,1)$, enabling smooth and stable modulation of the retrieved state contribution.

We integrate the retrieved state information with the attention outputs from the working sliding window and the non-evicted KV cache. The final attention output $\mathbf{O}_{t}$ is expressed as:
\begin{equation}
    \mathbf{O}_t = \text{Attn}_{local}(\mathbf{q}_t, \mathbf{K}_{local}, \mathbf{V}_{local}) +  \cdot \mathbf{O}_{\mathbf{S}}^{(t)}.
\end{equation}

\begin{table*}[t]
\centering
\caption{Performance comparison of different KV compression (The best results are highlighted in \textbf{bold}.)}
\label{tab:overall}
\small
\begin{tabular}{l|cccccccc}
\toprule \toprule
Method & TextVQA & DocVQA & MathVista & MMStar & MMMU & BLINK & MMCoQA & ALFRED  \\ \midrule
\multicolumn{9}{l}{\textbf{\textit{LLaVA-v1.5-7B}}} \\ \midrule
Full Cache   & 62.30 & 61.24 & 60.40 & 62.52 & 43.34 & 64.04 & 35.50 & 16.32 \\ \midrule
H2O          & 60.20 & 56.42 & \textbf{59.30} & 61.66 & 42.39 & 62.62 & 28.00 & 14.49 \\
SnapKV       & 61.00 & 58.52 & 59.10 & 62.13 & 42.33 & \textbf{63.14} & 30.50 & 14.57 \\
SAINT        & 60.00 & 57.84 & 58.60 & 61.84 & 42.28 & 62.64 & 28.00 & 14.64 \\
Meda         & 60.80 & 57.26 & 59.20 & 62.07 & 42.36 & 63.04 & 28.50 & 14.87 \\
LOOK-M       & 60.60 & 56.90 & 59.10 & 62.25 &  42.15 & 62.49 & 29.50 & 14.33 \\
RetentiveKV  & \textbf{61.20} & \textbf{59.05} & \textbf{59.30} & \textbf{62.42} &\textbf{42.45} & 63.12 & \textbf{31.50} & \textbf{15.82}\\ \midrule
\multicolumn{9}{l}{\textbf{\textit{Qwen3-VL-8B}}} \\ \midrule
Full Cache   & 62.00& 63.15 & 63.10 & 62.73 & 42.33 & 64.65 & 38.00 & 16.62  \\ \midrule
H2O          & 60.80 & 58.75 & 61.30 & 62.33 & 40.66 & \textbf{64.42} & 31.50 & \textbf{16.36} \\
SnapKV       & 61.30 & 60.12 & 62.20 & 62.36 & 41.38 & 64.31 & 32.50 & 16.12 \\
SAINT         & 60.60 & 60.44 & 60.50 & 62.38 & 41.22 & 63.82 & 32.00 & 16.15  \\
Meda         & 60.40 & 60.52 & 61.80 & 62.41 & 41.10 & 63.55 & 31.50 & 15.98 \\
LOOK-M       & 61.00 & 59.42 & 61.40 & 62.33 & 41.12 & 63.83 & 31.50 & 16.22 \\
RetentiveKV  & \textbf{61.40} & \textbf{62.21} & \textbf{62.60}& \textbf{62.58} & \textbf{41.57} & 64.22 & \textbf{33.10} & 16.34 \\ \midrule
\multicolumn{9}{l}{\textbf{\textit{Qwen3-VL-4B}}} \\ \midrule
Full Cache   & 57.70 & 59.97 & 50.40 & 60.66 & 42.11 & 64.95 & 32.50 & 15.98  \\ \midrule
H2O          & 55.00 & 56.03 & 48.00 & 60.36 &  41.05 & 63.55 & 28.00 & 14.67  \\
SnapKV       & 55.90 & 57.51 & \textbf{49.50} & 60.41 & 41.16 & 63.79 & \textbf{29.50} & 15.64 \\
SAINT        & 55.90 & 57.42 & 48.30 & 60.21 & 41.16 & 63.48 & 28.50 & 15.19 \\
Meda         & 55.60 & 57.51 & 48.80 & 60.38 & 41.38 & 63.62 & 28.50 & 15.26 \\
LOOK-M       & 56.10 & 56.64 & 48.50 & 60.25 & 41.32 & 63.21 & 28.00 & 15.32  \\
RetentiveKV  & \textbf{56.40} & \textbf{58.64} & 49.30 & \textbf{60.58} & \textbf{41.66} & \textbf{63.84} & \textbf{29.50} & \textbf{15.87}  \\ \bottomrule \bottomrule
\end{tabular}
\end{table*}
\section{Experiment}

\subsection{Experiment setting}
We evaluate the performance of RetentiveKV across three diverse and representative MLLM architectures: LLaVA-v1.5-7B, which serves as a widely recognized baseline for multimodal instruction tuning; Qwen3-VL-4B and Qwen3-VL-8B, representing state-of-the-art efficiency in general multimodal tasks; To verify the generalizability of our methods, we conduct a comprehensive evaluation across eight benchmarks covering multiple domains: MMMU \cite{mmmu} for expert-level reasoning, DocVQA \cite{docvqa} and TextVQA \cite{textvqa} for document and scene-text understanding, MathVista \cite{mathvista} for mathematical visual reasoning, MMStar \cite{mmstar} and BLINK for holistic perception, and MMCoQA \cite{milebench} and ALFRED\cite{milebench} for conversational and embodied AI tasks. We compare RetentiveKV against several competitive KV cache compression baselines, including importance-centric pruning methods: H2O \cite{h2o}, SnapKV \cite{snapkv}, and modality-aware compression techniques: LOOK-M \cite{lookm}, Meda \cite{meda}, SAINT \cite{SAINT}.

\subsection{Experiment Results}
Table~\ref{tab:overall} reports the overall performance comparison between RetentiveKV and state-of-the-art KV cache compression methods. The results demonstrate that RetentiveKV consistently outperforms prior KV eviction and compression strategies. In fine-grained visual perception tasks (\textit{e.g.,} DocVQA and TextVQA), RetentiveKV achieves +2.1 gains over the importance-centric baselines. These improvements stem from the entropy-guided state evolution, which preserves high-uncertainty visual tokens and enables their delayed reactivation, effectively mitigating premature eviction in importance-based methods. For long-context conversations and Needle-in-a-Haystack tasks such as MMCoQA and ALFRED, RetentiveKV improves over modality-aware baselines by +3.0 points. Unlike modality-aware methods that merely adjust compression ratios, RetentiveKV preserves deferred-critical visual information via continuous state evolution, enabling retrieval under long-horizon reasoning. The consistent performance gains observed across different model architectures (from LLaVA to Qwen3-VL) and parameter scales (4B to 8B) underscore the architectural agnosticism of our approach. 

\begin{table}[h]
\centering
\caption{Comparison of decoding latency and GPU memory usage for different cache budgets.}
\label{tab:latency}
\setlength{\tabcolsep}{3pt} 
\begin{tabular}{l|ccc}
\toprule
\textbf{Method} & \textbf{Budget} & \textbf{Latency} & \textbf{Memory} \\
\midrule
Full Cache & 100\% & 32.15 ms/token & 2.24 GiB \\
\midrule
\multirow{4}{*}{RetentiveKV} & 50\% & 27.84 ms/token & 1.18 GiB \\
 & 35\% & 24.42 ms/token & 0.84 GiB \\
 & 20\% & 21.42 ms/token & 0.46 GiB \\
 & 5\% & \textbf{18.42} ms/token & \textbf{0.23} GiB \\
\bottomrule
\end{tabular}
\end{table}

\begin{figure*}[t]
    \centering
    \includegraphics[width=\linewidth]{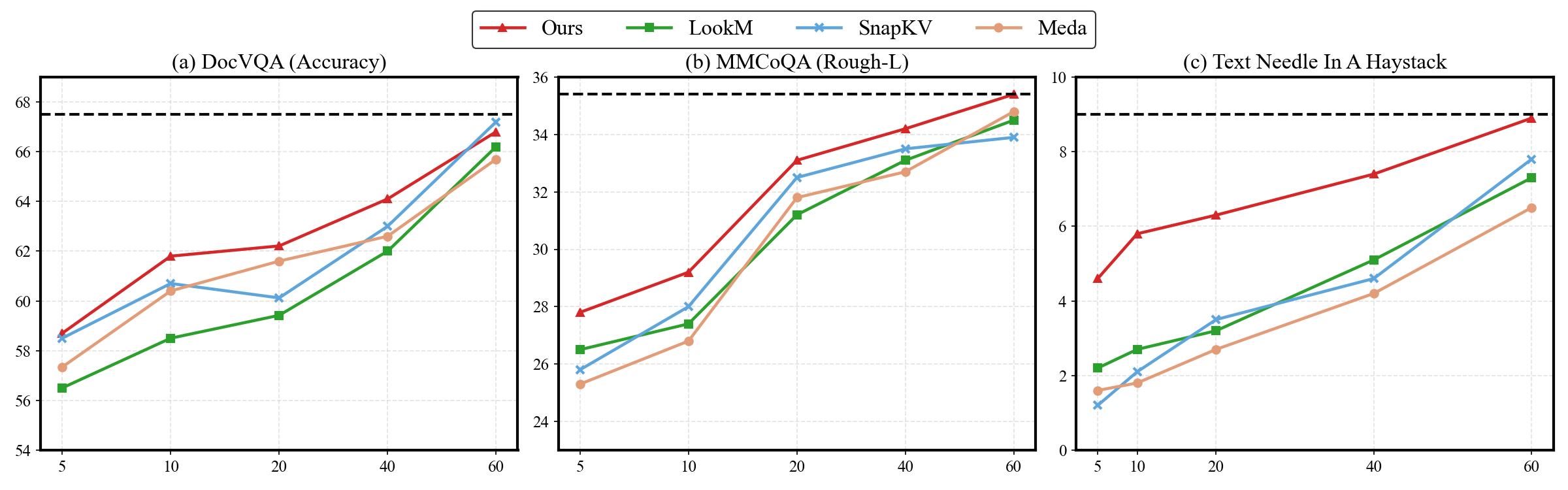}
    \caption{Comparison results for various cache budgets.}
    \label{fig:cache}
\end{figure*}

\subsection{Efficiency Analysis}
To investigate the computational efficiency of our approach, we evaluate the decoding latency and GPU memory footprint across varying compression ratios on a single NVIDIA A100 Tensor Core GPU. As presented in Table \ref{tab:latency}, RetentiveKV demonstrates superior resource efficiency compared to the Full Cache baseline. Specifically, under the most constrained budget of $5\%$, our method achieves a remarkable reduction in memory consumption, dropping from $2.24$ GiB to $0.23$ GiB. Simultaneously, it accelerates inference speed by $1.75\times$, reducing latency from $32.15$ ms/token to $18.42$ ms/token. Empirical observations reveal that GPU memory occupancy is directly proportional to the retained KV cache budget, suggesting that RetentiveKV effectively compresses context without introducing significant computational overheads.

\subsection{Ablation study}
To evaluate the individual contributions of our proposed components, we conduct extensive ablation studies. As detailed in Table \ref{tab:ablation}: 1) Removing state retention and Query-Conditioned Retrieval (QR) incurs the most severe performance degradation from $4.6\%$ to $5.4\%$. This mechanism allows the model to recall specific visual and textual cues that were previously evicted, mitigating the irreversible information loss inherent in discrete pruning. 2) Replacing the Modality-Agnostic state (MA) with a unified state space for all modalities degrades performance by $3.1\%$ on average. This disparity confirms the presence of cross-modal interference in unified structures. The specialized visual-dominant state prevents high-density textual sequences from overwhelming sparse visual cues. 3) Replacing the entropy-driven metric (EM) with a standard importance-centric metric leads to a degradation of $2.9\%$ on MMCoQA and $4.6\%$ on DocVQA. The marginal reduction suggests that entropy-driven selection is effective for long-context conversations.

\begin{table}[h]
\centering
\renewcommand{\arraystretch}{1.3}
\caption{Ablation study on key components of RetentiveKV.}
\label{tab:ablation}
\small
\setlength{\tabcolsep}{8pt} 
\begin{tabular}{ccc|cc}
\toprule
\textbf{EM} & \textbf{MA} & \textbf{QR} & \textbf{DocVQA} & \textbf{MMCoQA} \\ \midrule
\checkmark  &     \checkmark       &            & 58.70             & 60.91             \\
 \checkmark   &  &      \checkmark      & 60.20             & 61.55            \\
 & \checkmark &  \checkmark    & 59.80             &    61.38         \\ 
\checkmark  & \checkmark & \checkmark & \textbf{61.40}    & \textbf{64.22}    \\ \bottomrule
\end{tabular}
\end{table}

\subsection{Influence of Various Cache Budgets}
To evaluate the effectiveness of RetentiveKV under varying cache budgets, we conduct experiments on the \textit{Qwen3-VL-8B} model with cache budgets ranging from 5\% to 60\%. As shown in Figure \ref{fig:cache}, we observe that conventional importance-based eviction methods suffer from abrupt performance degradation when the cache budget is restricted to extremely low ratios (\textit{e.g.,} 5\%–10\%). This sharp decline confirms our hypothesis regarding Deferred Importance. In contrast, RetentiveKV exhibits robust performance stability. As illustrated in Figure \ref{fig:cache}(a) and (b), our method achieves a significant accuracy margin over the second-best baseline (SnapKV) on both DocVQA and MMCoQA tasks within the 10\%–20\% budget interval. The performance disparity is maximized in the "Text Needle In A Haystack" task (Figure \ref{fig:cache}(c)). At a constrained 5\% budget, RetentiveKV doubles the retrieval score of Meda and significantly surpasses SnapKV. This indicates that our entropy-driven state transition mechanism effectively compresses visual semantics into continuous states rather than discretely eliminating them, thereby preserving the spatial continuity and retrieving long-range dependencies required for later reasoning.

\section{Conclusion}
In this paper, we address the critical dilemma of Visual Continuity Collapse and Deferred Importance in multimodal KV cache compression. We introduce RetentiveKV, a framework that bridges the gap between discrete token pruning and continuous state modeling. Instead of viewing low-attention tokens as redundant noise, we reinterpret them through the information entropy, identifying those with high uncertainty as candidates for preservation within a continuously evolving state space. 
\newpage
\section{Limitation}
Despite the robust performance of RetentiveKV, two directions require further exploration. 1) Scaling Laws: Our current experiments validate efficacy on models up to 8B parameters. Future work will focus on investigating the behavior of entropy-driven retention in massive-scale MLLMs (>30B), particularly to determine if the "deferred importance" phenomenon intensifies as model capacity grows. 2) Towards Omni-modal Perception: While this paper addresses visual-linguistic challenges, the "continuous evolution" paradigm of RetentiveKV is inherently agnostic to data modality. As the field advances towards Omni-modal MLLMs, we see significant potential in applying RetentiveKV to real-time audio and video. Leveraging the continuous nature of SSMs to unify memory management across diverse modalities stands as a compelling direction for our future research.

\section{Acknowledgements}
This work was supported by the Key Research and Development Program of Zhejiang Province(No. 2025C01026), and the National Natural Science Foundation of China (No. 62402429, U24A20326, 62441236). This work was also partially supported by the Ningbo Yongjiang Talent Introduction Programme (2023A-397-G) and Young Elite Scientists Sponsorship Program by CAST (2024QNRC001). The author gratefully acknowledges the support of Zhejiang University Education Foundation Qizhen Scholar Foundation.
\bibliography{custom}

\end{document}